\documentclass{article} 
\usepackage{iclr2025_conference,times}

\usepackage{amsmath,amsfonts,bm}

\def\eqref#1{equation~\ref{#1}}

\def\1{\bm{1}}

\DeclareMathAlphabet{\mathsfit}{\encodingdefault}{\sfdefault}{m}{sl}
\SetMathAlphabet{\mathsfit}{bold}{\encodingdefault}{\sfdefault}{bx}{n}

\usepackage{hyperref}
\usepackage{url}
\usepackage{booktabs}
\usepackage{tabularx}
\usepackage{ragged2e}
\usepackage{subcaption}
\usepackage{caption}
\usepackage{latexsym}
\usepackage{amsmath} 
\usepackage{amsfonts}
\usepackage{amssymb}
\usepackage{subcaption}
\usepackage{graphicx}

\newcommand{\modelname}[0]{CoPRIS}
\newcommand{\tablesize}{\fontsize{8.2pt}{9.5pt}\selectfont}
\title{CoPRIS: Efficient and Stable Reinforcement Learning via Concurrency-Controlled Partial Rollout with Importance Sampling}

\iclrfinalcopy
\author{
Zekai Qu$^{1}$,
Yinxu Pan$^{1}$,
Ao Sun$^{1}$,
Chaojun Xiao$^{2}$\thanks{~Corresponding authors.},
Xu Han$^{2*}$ \\
$^{1}$OpenBMB \quad
$^{2}$Tsinghua University \\
\texttt{\{zekaiqu23\}@gmail.com, \{xcj,han-xu\}@tsinghua.edu.cn}
}

\begin{document}

\maketitle

\begin{abstract}
Reinforcement learning (RL) post-training has become a trending paradigm for enhancing the capabilities of large language models (LLMs).
Most existing RL systems for LLMs operate in a fully synchronous manner, where training must wait for the rollout of an entire batch to complete. This design leads to severe inefficiencies, as extremely long trajectories can stall the entire rollout process and leave many GPUs idle. To address this issue, we propose Concurrency-Controlled Partial Rollout with Importance Sampling (\modelname{}), which mitigates long-tail inefficiencies by maintaining a fixed number of concurrent rollouts, early-terminating once sufficient samples are collected, and reusing unfinished trajectories in subsequent rollouts. To mitigate the impact of off-policy trajectories, we introduce Cross-stage Importance Sampling Correction, which concatenates buffered log probabilities from the previous policy with those recomputed under the current policy for importance sampling correction. Experiments on challenging mathematical reasoning benchmarks show that \modelname{} achieves up to 1.94× faster training while maintaining comparable or superior performance to synchronous RL systems. The code of \modelname{} is available at \href{https://github.com/777pomingzi/CoPRIS}{https://github.com/777pomingzi/CoPRIS}.
\end{abstract}

\section{Introduction}
Reinforcement learning (RL) has emerged as a new scaling paradigm for enhancing large language models’ (LLMs) reasoning capabilities on complex tasks. Advanced models such as DeepSeek-R1~\citep{deepseekr1} and o1~\citep{openai-o1} have demonstrated the remarkable effectiveness of RL in improving LLM performance across diverse domains, including mathematical reasoning~\citep{math,deepscaler2025}, code generation~\citep{livecodebench,deepcoder2025}, scientific problem-solving~\citep{gpqa,last-exam,general}, and tool use~\citep{tau-bench}. The RL post-training workflow typically comprises three key stages: rollout, reward, and training. In the rollout stage, the actor LLM generates responses to a batch of input prompts. During the reward stage, these responses are evaluated through various reward mechanisms. Finally, during training, the obtained rewards are used to compute the loss, providing optimization signals for updating the model parameters.

Due to the inherent temporal dependencies in the RL post-training pipeline, most large-scale RL systems adopt a fully synchronous design~\citep{trl,verl,roll,OpenRLHF}. In such a synchronous setup, the training stage must wait until all batch trajectories are completed before starting. However, this design introduces significant computational bubbles. During rollout, response lengths exhibit significant variance across prompts, resulting in a pronounced long-tail distribution. As a result, some GPUs that complete their assigned rollout tasks early are forced to idle while waiting for a few exceptionally long responses to finish, causing inefficient utilization of computational resources.

To improve training efficiency, asynchronous RL training systems have received increasing attention~\citep{areal,history}. 
These systems can be broadly categorized into two paradigms.
In the \textit{rollout-training separation} paradigm, computational resources are allocated to rollout and training workers: rollout workers continuously generate trajectories into a shared buffer, where training workers retrieve trajectories for updates.
This design eliminates idle time but introduces a critical challenge in balancing generation and consumption rates: when rollout workers generate trajectories too quickly, the buffer accumulates stale off-policy data; conversely, when generation is too slow, training workers are starved of samples.
Alternatively, the \textit{rollout-training integration} paradigm, exemplified by Kimi-K1.5~\citep{kimi1.5}, employs partial rollout strategies. It over-generates trajectories in parallel and halts generation once a full batch is collected, caching incomplete responses for subsequent reuse.
While this approach eliminates computational idle time, it faces two critical challenges:
(1)~\textbf{Efficiency and Stability degradation from uncontrolled concurrency.} Without proper concurrency control, the system may allocate excessive prompts to a single GPU, exceeding its memory capacity. This triggers the key-value recomputation mechanism, introducing substantial computational overhead that degrades overall throughput. Furthermore, excessive concurrency also increases off-policy trajectories, amplifying distributional drift and training instability.
(2)~\textbf{Training instability from off-policy trajectories.} The asynchronous nature inevitably produces trajectories partially generated by earlier policy. Directly training on these off-policy trajectories leads to a distribution mismatch that degrades training stability and ultimately impairs model performance.

To address these challenges, we propose Concurrency-Controlled Partial Rollout with Importance Sampling~(\modelname{}) — a computationally efficient framework designed to accelerate RL training without compromising model performance. 
Building upon the naive partial rollout paradigm, \modelname{} introduces two key mechanisms to overcome the aforementioned limitations.
First, we introduce \textbf{Concurrency-Controlled Generation} to regulate the number of concurrent rollouts. This mechanism prevents memory overflow and the associated recomputation overhead, while simultaneously mitigating excessive off-policy trajectories by limiting excessive trajectory generation.
Second, we incorporate \textbf{Cross-stage Importance Sampling Correction} to mitigate the distribution mismatch introduced by off-policy trajectories. When training with buffered trajectories, each token preserves the log-probability computed under the policy model corresponding to the rollout stage in which the token is generated. These stored log probabilities are subsequently concatenated with those computed under the current policy to perform importance sampling correction. Through these mechanisms, \modelname{} eliminates the computational idle time caused by the long-tail problem during rollout and prevents off-policy trajectories from adversely affecting training stability. 

To verify the effectiveness of \modelname{}, we deploy it in a widely-used open-sourced RL framework, veRL~\citep{verl}, and evaluate it on $5$ challenging mathematical benchmarks using models of three different parameter scales. Under strictly controlled and comparable conditions, \modelname{} achieves an end-to-end speedup of $1.58\times$–$1.94\times$ while maintaining comparable or even superior performance to the fully synchronous RL method. Furthermore, our comprehensive ablation studies demonstrate that, compared to the naive partial rollout approach, \modelname{} consistently delivers stable improvements in both training efficiency and model performance through its concurrency control and importance sampling mechanisms.

\section{Related Work}

\subsection{Synchronous RL}
In a synchronous RL framework, a batch of prompts is sent to the inference engine to generate trajectories, and only after all trajectories are completed is it passed to the training engine. This paradigm arises from the inherent on-policy nature of RL and is widely adopted in many RL training frameworks~\citep{trl,verl,OpenRLHF,roll}.
It ensures that all trajectories are directly sampled from the current policy model, thereby guaranteeing training stability and facilitating convergence. However, this approach suffers from the long-tail problem, as it requires waiting for a few exceptionally long trajectories to finish~\citep{kimi}, leading to idle computation and resource waste.

\subsection{Asynchronous RL}
To address the inefficiencies of synchronous RL, researchers have begun exploring asynchronous RL systems. In an asynchronous RL framework, the inference engine continuously generates trajectories and places completed ones into a buffer, while the training engine independently retrieves training data from this buffer for parameter updates. This decoupling maximizes hardware utilization and markedly improves throughput~\citep{tba,async-rlhf,topr}. 
These systems typically adopt \textit{rollout-training separation} paradigm~\citep{areal,AsyncFlow,streamrl,llamarl}, where GPU devices are divided into rollout and training workers. Rollout workers generate trajectories into a shared buffer for training workers to retrieve asynchronously. While maximizing hardware utilization, this design faces challenges in balancing generation and consumption rates.
An alternative approach employs \textit{rollout-training integration} via partial rollout~\citep{kimi1.5}: over-generating trajectories in parallel and caching incomplete responses for reuse. However, uncontrolled concurrency can cause memory pressure and recomputation overhead, while excessive off-policy trajectories introduce distribution mismatch affecting training stability.
To address these issues, we introduce Concurrency-Controlled Generation and Cross-stage Importance Sampling Correction, which regulate off-policy trajectories and mitigate off-policy bias by applying importance sampling correction over concatenated log probabilities, ensuring stable and efficient training.

\section{Background}
\subsection{Preliminary about RL Training}
We cast reinforcement learning (RL) training process into the Markov Decision Process (MDP) framework~\citep{mdp}, 
characterized by the tuple $\langle S, A, r, P, \gamma, H \rangle$. 
Here, $S$ denotes the state space, $A$ the action space, $P$ the transition model, 
$r: S \times A \to \mathbb{R}$ the reward function, $\gamma$ the discount factor, 
and $H$ the horizon. 
The LLM adopts a parameterized policy 
$\pi_{\theta}: S \to A$, where each action $a_t \in A$ corresponds to a token in the vocabulary. 
The state $s_t \in S$ is defined as a question $s_1 = q$ concatenated with the sequence of 
previously generated tokens $(a_1, \dots, a_{t-1})$, with deterministic updates 
$s_{t+1} = \text{concat}(s_t, a_t)$. 
Given a question distribution $D$, our goal is to maximize the objective:
\begin{equation}
J(\theta) 
= \mathbb{E}_{q \sim D,\, a_t \sim \pi_\theta(\cdot \mid q, a_{<t})} 
\left[ \sum_{t=1}^{H} \gamma^{t-1} r(s_t, a_t) \right].
\end{equation}

Following~\citet{deepseekr1}, we employ a rule-based reward function that only provides reward on the final action, and set $\gamma = 1$. 
We optimize this objective using \emph{Group Relative Policy Optimization} (GRPO)~\citep{grpo}, which computes advantages by intra-group comparison:
\begin{align}
J_{\mathrm{GRPO}}(\theta)
&= \mathbb{E}_{\,q\sim D,\; a_t \sim \pi_{\theta_{\mathrm{old}}}(\cdot \mid q,a_{<t})} \left[
\frac{1}{G} \sum_{i=1}^{G} \frac{1}{|o_i|} \sum_{t=1}^{|o_i|} \mathcal{L}_{i,t}(\theta)
- \beta\, D_{\mathrm{KL}}\!\left(\pi_{\theta} \,\|\, \pi_{\mathrm{ref}}\right)
\right],
\label{eq:grpo_combined}
\end{align}

\begin{align}
\mathcal{L}_{i,t}(\theta) 
= & \;\min \Big( r_{i,t}(\theta)\,\hat{A}_{i,t},  \operatorname{clip}\!\big(r_{i,t}(\theta),\, 1-\epsilon,\, 1+\epsilon \big)\,\hat{A}_{i,t} \Big),
\label{eq:grpo_step}
\end{align}
where
\begin{equation}
r_{i,t}(\theta) = 
\frac{\pi_{\theta}(o_{i,t} \mid q, o_{i,<t})}
{\pi_{\theta_{\mathrm{old}}}(o_{i,t} \mid q, o_{i,<t})},
\label{eq:ratio}
\end{equation}

\begin{equation}
\hat{A}_{i,t} = 
\frac{R_i - \operatorname{mean}\!\left(\{R_j\}_{j=1}^{G}\right)}
     {\operatorname{std}\!\left(\{R_j\}_{j=1}^{G}\right)}.
\label{eq:adv_norm}
\end{equation}
where $\beta$ is a hyperparameter that controls the KL regularization strength, $G$ denotes the number of rollouts per question, and $o_{i,t}$ is the $t$-th token in the $i$-th output sequence $o_i$ within the group.
\subsection{Long-tail Problem of RL Post-Training}
\label{long-tail problem}
\begin{figure*}[htbp]
    \centering
    \begin{subfigure}[b]{0.45\textwidth}
        \centering
        \includegraphics[width=\textwidth]{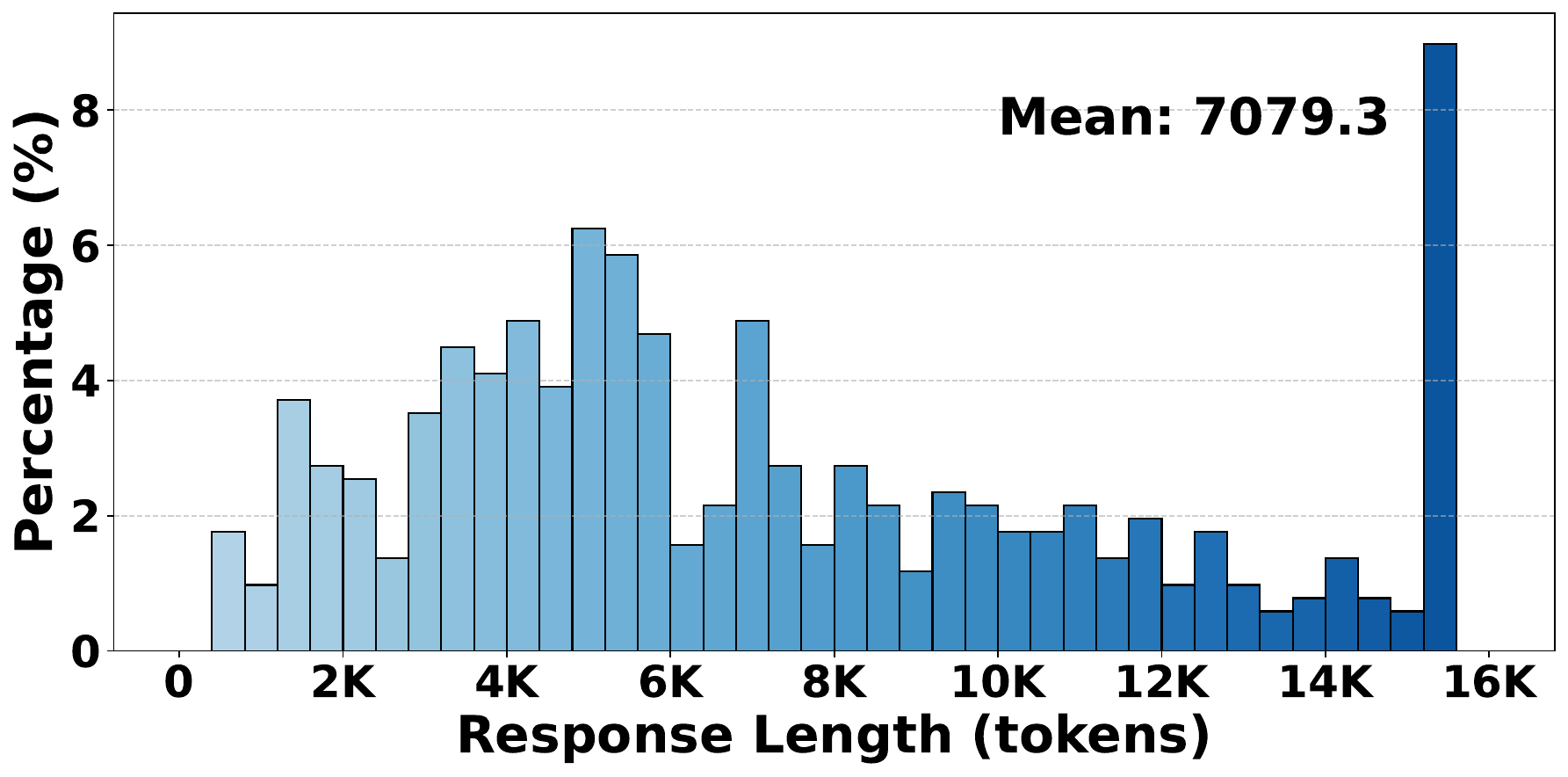}
        \caption{Response length distribution}
        \label{fig:subfig1}
    \end{subfigure}
    \hfill
    \begin{subfigure}[b]{0.45\textwidth}
        \centering
        \includegraphics[width=\textwidth]{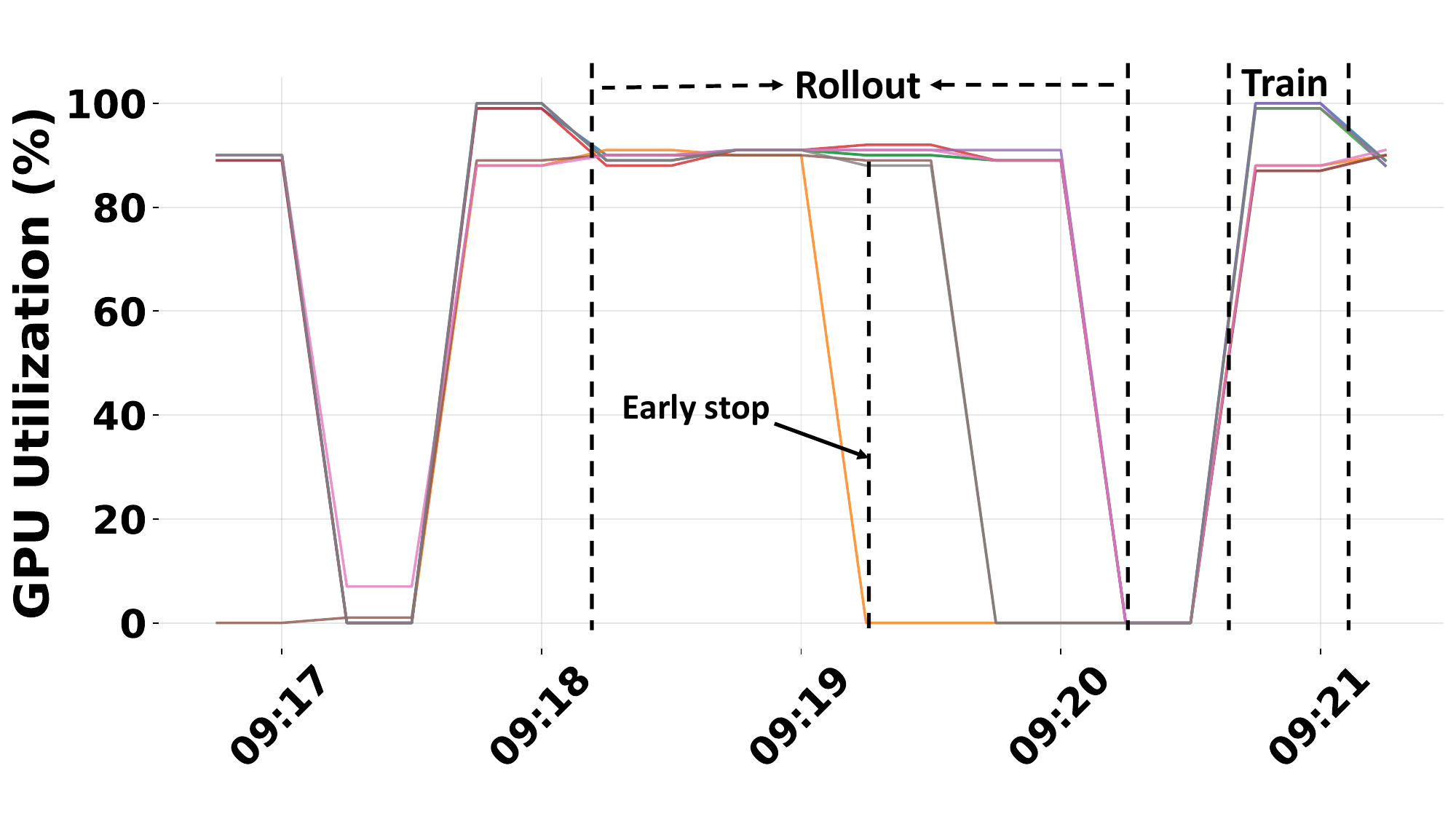}
        \caption{GPU utilization}
        \label{fig:subfig2}
    \end{subfigure}
    \caption{RL training traces for a single training step of DeepSeek-R1-Distill-Qwen-7B. Experiments are conducted on DeepScaleR using 32 NVIDIA H800 GPUs with a maximum context length of 16k tokens. The wall-clock GPU utilization trajectories of 8 GPUs are visualized.}

    \label{fig:twofigs}
\end{figure*}

In RL post-training, the rollout stage remains the dominant bottleneck. Response lengths vary substantially across inputs, yielding a long-tailed distribution: within a batch, most samples must wait for a small minority of trajectories with exceptionally long responses to finish. This induces computation bubbles and leaves some workers idle. The within-batch long-tail is evident in Figure~\ref{fig:subfig1}, where response lengths differ widely across trajectories. Consequently, some rollout workers finish early and sit idle until stragglers complete, producing pronounced utilization dips  (Figure~\ref{fig:subfig2}).

\section{Methodology}
\begin{figure*}[t]
    \centering
    \includegraphics[width=1\textwidth]{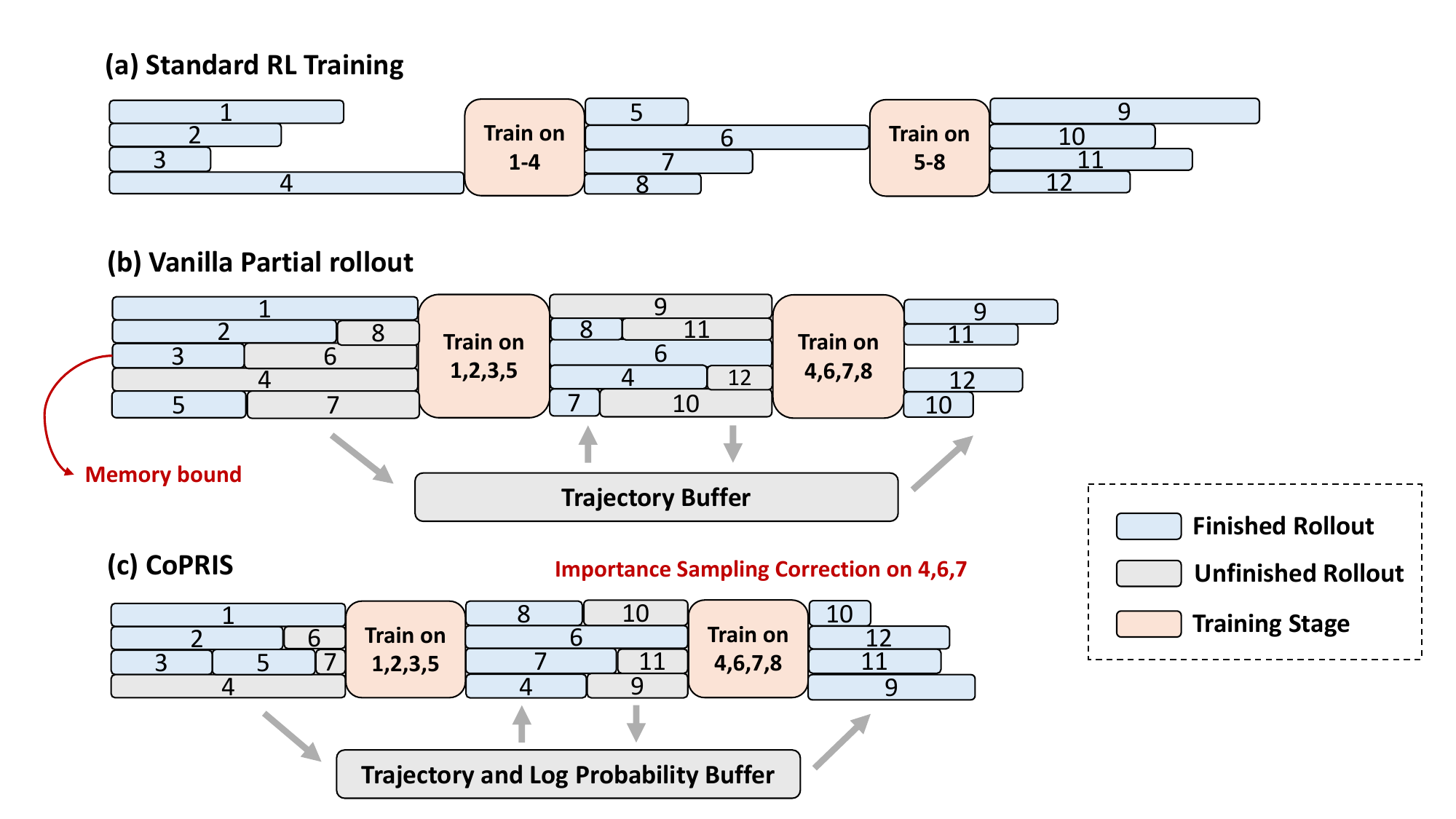}
    \caption{Illustration of rollout management in \modelname{}. The number of concurrent generations remains constant and is independent of the training batch size. The buffer stores incomplete trajectories together with their corresponding log probabilities under the policy, enabling subsequent continuation and importance sampling correction.}
    \label{fig:partial}
\end{figure*}
To address the long-tail problem discussed in Section~\ref{long-tail problem}, we draw inspiration from the partial rollout mechanism introduced in Kimi-K1.5~\citep{kimi1.5}. Building upon this foundation, we propose \textbf{Concurrency-Controlled Generation} and \textbf{Cross-stage Importance Sampling Correction} to mitigate the adverse effects of off-policy trajectories and enhance computational efficiency.
We integrate these components into our framework, termed \modelname{}, as illustrated in Figure~\ref{fig:partial}. Its detailed workflow is described as follows:

\textbf{Concurrency-Controlled Generation.} 
During rollout, \modelname{} maintains a fixed number of rollout requests \(N'\) within the inference engines. Each request corresponds to an input–trajectory pair. Whenever a trajectory finishes generation, a new request is immediately dispatched, ensuring that \(N'\) requests remain active and that computational resources are fully utilized. Moreover, by adjusting the concurrency level, we can control the number of off-policy trajectories and reduce the additional recomputation overhead that may arise under excessively high concurrency.

\textbf{Early Termination.} 
The rollout process continues until the required batch size \(B\) for a single training step is reached. 
Once \(N\) trajectories have been collected for each of the \(B\) inputs (where \(N\) denotes the number of rollouts per input), the inference engines immediately halt generation.

\textbf{Buffering of Partial Trajectories.} 
The trajectories that are partially generated when early termination occurs are not discarded. 
Instead, \modelname{} stores them in a buffer, along with their log probabilities computed under corresponding policy models. 
Formally, for a trajectory 
$\tau_i = \{o_{i,1}, o_{i,2}, \dots, o_{i,t}\}$ 
conditioned on the input query $q$, the corresponding log-probability is given as
\begin{equation}
\mathbf{L}_i = 
\operatorname{concat}\!\big(
\mathbf{L}_i^{(1)},\,
\mathbf{L}_i^{(2)},\,
\dots,\,
\mathbf{L}_i^{(K)}
\big),
\label{eq:logprob_concat}
\end{equation}
where $\mathbf{L}_i^{(k)}$ represents the log probabilities 
of the subsequence generated by policy $\pi_{\theta^{(k)}}$ at stage $k$.
The operator $\operatorname{concat}(\cdot)$ denotes the concatenation along the token dimension. 
    
Besides, the buffer $\mathcal{B}$ stores not only unfinished trajectories but also 
completed ones whose corresponding input groups have not yet finished generation.
Formally,
\begin{equation}
\mathcal{B} = 
\left\{
\left(\tau_i, \mathbf{L}_i\right)
\,\middle|\,
i \in \mathcal{I}_{\text{active}}
\right\},
\label{eq:buffer}
\end{equation}
where $\mathcal{I}_{\text{active}}$ indexes all trajectories that are still associated 
with active rollout groups. 
Both unfinished trajectories and those already completed but belonging to still-active groups 
retain their log-probability sequences in the buffer. 
Unfinished trajectories have their sequences extended in subsequent rollouts, while completed trajectories remain unchanged; both are used for importance sampling correction in future training steps.

\textbf{Cross-stage Importance Sampling Correction.} 
During training, \modelname{} performs importance sampling correction
using the cached log-probabilities stored in the buffer $\mathcal{B}$.
For each buffered trajectory 
$\tau_i = \{o_{i,1}, o_{i,2}, \dots, o_{i,t}\}$ 
with stored log-probabilities $\mathbf{L}_i$,
\modelname{} recomputes the corresponding log-probabilities under the current policy
$\pi_{\theta}$ and applies importance correction.

Following the definition in Eq.~\eqref{eq:ratio}, the importance ratio for each token is given by:
\begin{equation}
r_{i,t}(\theta) = \frac{\pi_{\theta}(o_{i,t} \mid q, o_{i,<t})}{\pi_{\theta_{\mathrm{old}}}(o_{i,t} \mid q, o_{i,<t})}
=
\exp\!\big(
\mathbf{L}_{i,t}^{(\theta)} - \mathbf{L}_{i,t}
\big),
\quad
\label{eq:token_importance}
\end{equation}

These per-token ratios $\{r_{i,t}(\theta)\}$ are used to reweight objectives during policy optimization, allowing \modelname{} to effectively utilize off-policy trajectories.

\textbf{Prioritized Resumption.} 
In the next rollout stage, the inference engine prioritizes retrieving the buffered partial trajectories and continues their generation.

By buffering partially generated responses of long trajectories and reusing them in subsequent rollouts while maintaining a fixed number of concurrent requests, \modelname{} fully utilizes computational resources, thereby improving system throughput and accelerating training. Moreover, by appropriately controlling the concurrency level, \modelname{} achieves a balance between training effectiveness and efficiency—since excessive concurrency not only reduces computational efficiency due to redundant inference recomputation, but also increases the number of off-policy trajectories and the cost of log-probability calculations. Finally, through Cross-stage Importance Sampling Correction, \modelname{} ensures both the stability and effectiveness of training across stages.

\section{Experiments}

To verify the effectiveness of \modelname{}, our experiments cover: (1) Performance of \modelname{} when applied to state-of-the-art open-source frameworks across different models. (2) The acceleration achieved by \modelname{} under varying model sizes and context lengths. (3) Ablation studies on the proposed Concurrency-Controlled Generation and Cross-stage Importance Sampling Correction.

\subsection{Experimental Setup}
\label{setup}

We evaluate \modelname{} on a suite of challenging mathematical tasks, including AIME 2024~\citep{AIME2024}, AIME 2025, AMC~\citep{numinamath}, Minerva Math ~\citep{lewkowycz2022solving}, and OlympiadBench~\citep{olympiadbench}. We use DeepSeek-R1-Distill-Qwen (1.5B and 7B)~\citep{qwen2.5,deepseekr1} and Qwen3-8B~\citep{qwen3technicalreport} as the base models.

For training, we use Group Relative Policy Optimization (GRPO)~\citep{grpo} as the reinforcement learning algorithm and the DeepScaleR-Preview-Dataset~\citep{deepscaler2025} as the training dataset, with detailed hyperparameter settings listed in Table~\ref{tab:setup}. All models are trained for 1,000 steps, with DeepSeek-R1-Distill-Qwen-1.5B using 16 A800 GPUs, and DeepSeek-R1-Distill-Qwen-7B and Qwen3-8B using 32 H800 GPUs.

To ensure a rigorous evaluation of \modelname{} in terms of both training efficiency and effectiveness, we implement it on the latest release of the veRL repository (main branch, August 18, 2025)~\citep{verl}. Furthermore, to guarantee fairness in comparison, asynchronous rewards are applied to both the baseline and \modelname{}, thereby enabling a reliable assessment of \modelname{}’s effectiveness.

\begin{table*}[t]
\centering
\tablesize
\begin{tabular}{l c c c c c c c c} 
\toprule 
Model & AIME24 & AIME25 & AMC &  MinervaMath & OlympiadBench & Average & Training Hours $\downarrow$ \\
\midrule
\multicolumn{8}{c}{\textbf{Distill-Qwen-1.5B}} \\ \midrule
Basemodel & 29.38 & 23.96 &58.92 &22.73 &39.41   &34.88 & - \\
veRL & 32.60 & 25.31 &63.97 &23.81 &43.07  &37.75 & 54.05 \\
\modelname{} & \textbf{34.48} & \textbf{28.33} &\textbf{65.89} &\textbf{24.43} &\textbf{43.60}  &\textbf{39.36} & \textbf{34.20 (1.58 $\times$)} \\
\midrule
\multicolumn{8}{c}{\textbf{Distill-Qwen-7B}} \\ \midrule
Basemodel &52.29  & 36.77 &78.13 &30.57 &51.30   &49.81 & - \\
veRL & 54.37 & 42.40 &\textbf{82.34} &31.00 & 53.19  &52.66 & 43.57 \\
\modelname{} &\textbf{55.73}  & \textbf{43.85} &82.00  &\textbf{32.49} &\textbf{54.34} &\textbf{53.68} & \textbf{22.44 (1.94 $\times$)} \\
\midrule
\multicolumn{8}{c}{\textbf{Qwen3-8B}} \\ \midrule
Basemodel &59.06  & 47.50 &78.13 &35.86 &55.16  & 55.14& - \\
veRL &\textbf{69.79}  & \textbf{57.41} &84.45 &\textbf{36.12}  &58.64 &\textbf{61.28} &54.42  \\
\modelname{} &67.71  & 55.52 &\textbf{85.54} &35.95 &\textbf{59.65}  &60.87 & \textbf{31.16 (1.75 $\times$)} \\
\bottomrule 
\end{tabular}
\caption{End-to-End Performance Comparison. Our evaluation is conducted using the official evaluation interface provided by veRL, reporting the average pass@1 accuracy.}
\label{main_result}
\end{table*}

\subsection{End-to-End Comparison}

We evaluate the end-to-end performance of \modelname{} across three model scales using the experimental setup described in Section~\ref{setup}. Table~\ref{main_result} presents our experimental findings. The results demonstrate that \modelname{} consistently achieves substantial speedups across all three model scales, reducing wall-clock training time by 1.58$\times$–1.94$\times$. Beyond efficiency gains, \modelname{} maintains comparable or superior performance relative to the baseline, demonstrating notable improvements on Distill-Qwen-1.5B and Distill-Qwen-7B with average score increases of 1.61 and 1.02 points, respectively. On Qwen3-8B, \modelname{} exhibits marginally lower performance, though this minor trade-off is accompanied by substantial training acceleration of 1.75$\times$. These results underscore the effectiveness of our asynchronous training paradigm in substantially reducing training duration without compromising model performance.

We hypothesize that the performance variations across models—improvements on Distill-Qwen-1.5B and 7B versus marginal degradation on Qwen3-8B—arise from the interaction between off-policy exploration and initial model entropy. Our observations reveal that Qwen3-8B exhibits lower initial entropy than Distill-Qwen-1.5B and 7B, indicating varying degrees of policy confidence at initialization. Models with higher initial entropy naturally explore more broadly, making them more compatible with \modelname{}'s off-policy trajectories, while lower-entropy models may experience excessive exploration that induces distributional shift. This is corroborated by entropy dynamics throughout training: \modelname{} consistently maintains higher entropy than the baseline across all models. Notably, Distill-Qwen-1.5B and 7B exhibit characteristic rise-then-fall entropy trajectories, successfully leveraging off-policy exploration before converging to stable policies. In contrast, Qwen3-8B demonstrates monotonically increasing entropy without convergence stabilization, resulting in sustained exploration and marginal performance degradation.

These findings suggest that \modelname{} requires calibration of the off-policy trajectory proportion based on initial model entropy. For higher-entropy models, \modelname{} yields both substantial speedup and enhanced performance. For lower-entropy models, while significant acceleration remains achievable, more conservative tuning of the off-policy ratio is advisable to balance exploration benefits against potential instability.

\begin{figure*}[t] 
    \centering
    \begin{subfigure}[t]{0.48\linewidth}
        \centering
        \includegraphics[width=\linewidth]{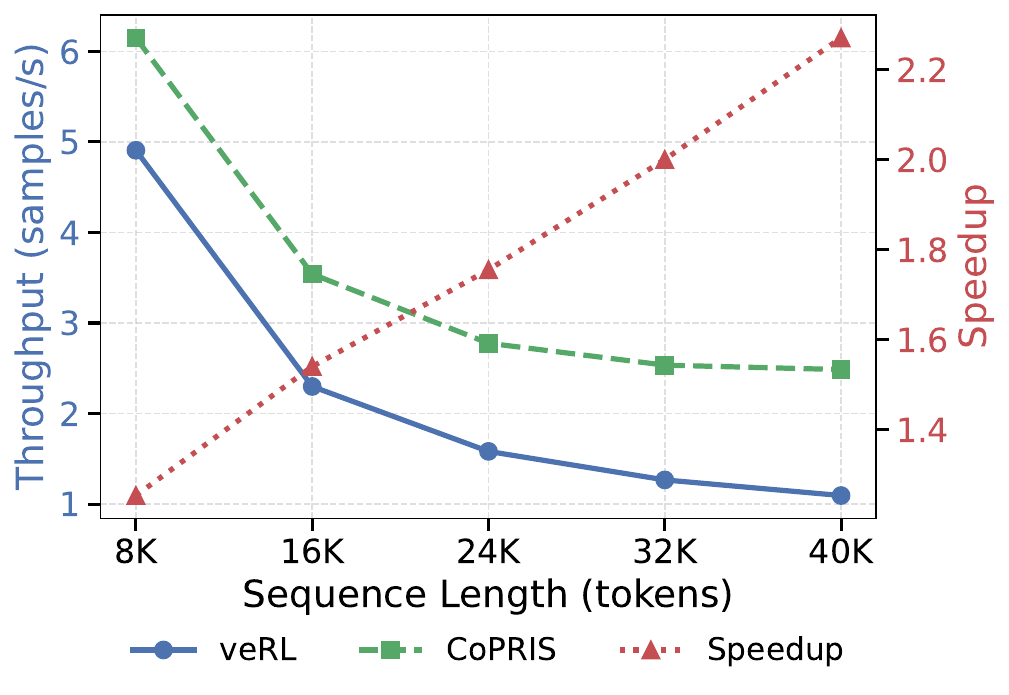}
        \caption{Scaling with context length.}
    \end{subfigure}
    \hfill
    \begin{subfigure}[t]{0.48\linewidth}
        \centering
        \includegraphics[width=\linewidth]{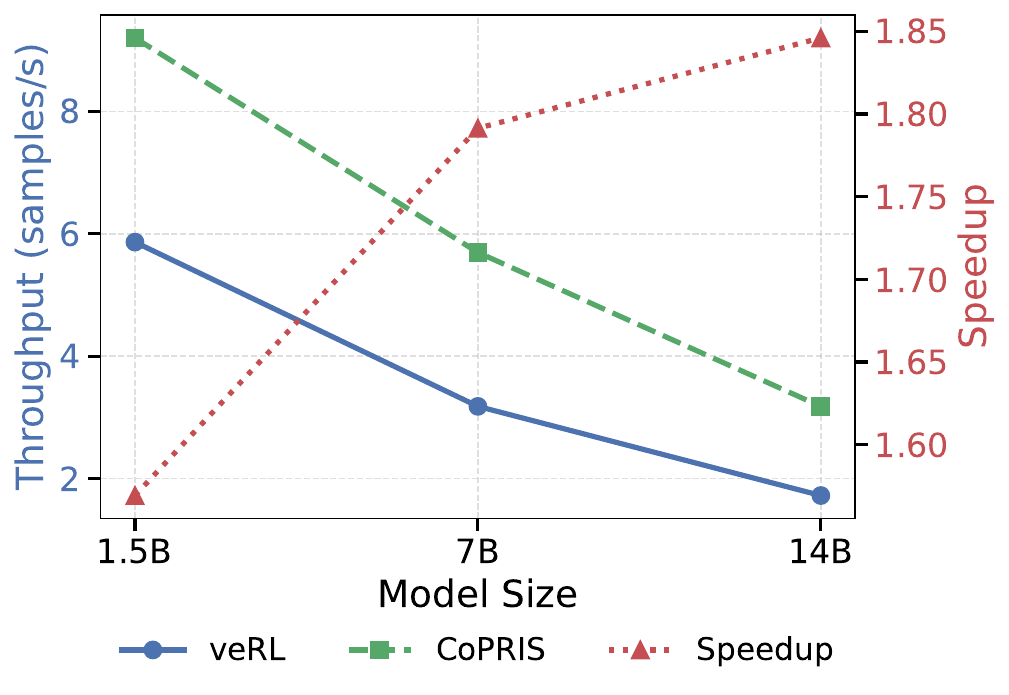}
        \caption{Scaling with model size.}
    \end{subfigure}

    \caption{Scalability of \modelname{}. Throughput and speedup comparison with veRL under different context lengths and model sizes.}
    \label{fig:pris_scalability}
\end{figure*}

\subsection{Scalability Analysis of \modelname{}}
We evaluate the scalability of \modelname{} against veRL across two dimensions: context length and model size. 
Experiments are conducted on 32 H800 GPUs using the DeepScaleR-Preview Dataset for 50 steps. 
We report the average effective throughput over these 50 steps, defined as the rate at which generated training samples are consumed during updates, with all other configurations remaining consistent with Section~\ref{setup}. 
For the context-length scaling experiments, we employ Qwen3-8B as the base model due to its extended context window. 
For the model-size scaling experiments, we utilize the DeepSeek-R1-Distill-Qwen family with parameter sizes of 1.5B, 7B, and 14B. 

The results are presented in Figure~\ref{fig:pris_scalability}. 
From the experimental results, we observe that:
(1) \textit{Context length scaling}: As context length increases from 8K to 40K, \modelname{} consistently outperforms veRL with progressively larger speedups. Specifically, the speedup increases from 1.27$\times$ at 8K tokens to 2.26$\times$ at 40K tokens, demonstrating a near-linear growth trend. This substantial improvement validates the effectiveness of our trajectory reuse strategy in mitigating the long-tail problem during the rollout stage. The results highlight the significant potential of \modelname{} in long chain-of-thoughts RL training scenarios where rollout completion times become increasingly heterogeneous.
(2) \textit{Model size scaling}: Across different model sizes (1.5B, 7B, 
and 14B parameters), \modelname{} consistently achieves higher throughput than veRL, with speedups ranging from 1.57$\times$ to 1.85$\times$. Notably, these results were obtained using a fixed concurrency and computational resource configuration across all model sizes, which may not be optimal. For larger models, a lower concurrency level can already fully saturate the available computational resources. Maintaining the same concurrency level can trigger memory-induced recomputation overhead and generate redundant off-policy trajectories that require re-prefilling and log-probability calculation for importance sampling correction. These combined effects limit the full potential of \modelname{} at larger scales. Therefore, dynamically adjusting the concurrency limit based on model size and computational resources would likely yield even greater performance improvements for \modelname{}.

\begin{table*}[t]
\centering
\renewcommand{\arraystretch}{1.15}
\setlength{\tabcolsep}{9pt}
\tablesize
\begin{tabular}{cccccc}
\toprule
\textbf{Concurrency} & \textbf{AIME24 $\uparrow$} & \textbf{AIME25 $\uparrow$} & \textbf{Step/s $\downarrow$} & \textbf{Rollout/s $\downarrow$} & \textbf{Cal logprob/s $\downarrow$} \\
\midrule
\textbf{Naive Partial Rollout (1536)}  & 34.38 & 28.23 &126.81  &77.09  &23.79   \\
\textbf{512}  & 34.06 & 27.83 &139.08  &97.44  &  \textbf{15.96} \\
\textbf{1024} &\textbf{34.48}  &\textbf{28.33}  &\textbf{123.12}  &\textbf{75.43}  & 22.15\\
\textbf{1536} & 32.50 & 26.25 &144.00  &88.15  &  29.21 \\
\textbf{2048} & 33.23 & 25.83 &161.06  & 95.44 &  37.30 \\
\bottomrule
\end{tabular}
\caption{Impact of different concurrency levels on \modelname{}'s performance and efficiency.}
\label{tab:concurrency}
\end{table*}

\begin{figure*}[t] 
\centering
\setlength{\abovecaptionskip}{2pt}
\setlength{\belowcaptionskip}{-8pt}

\begin{subfigure}[t]{0.48\textwidth}
    \centering
    \includegraphics[width=\linewidth]{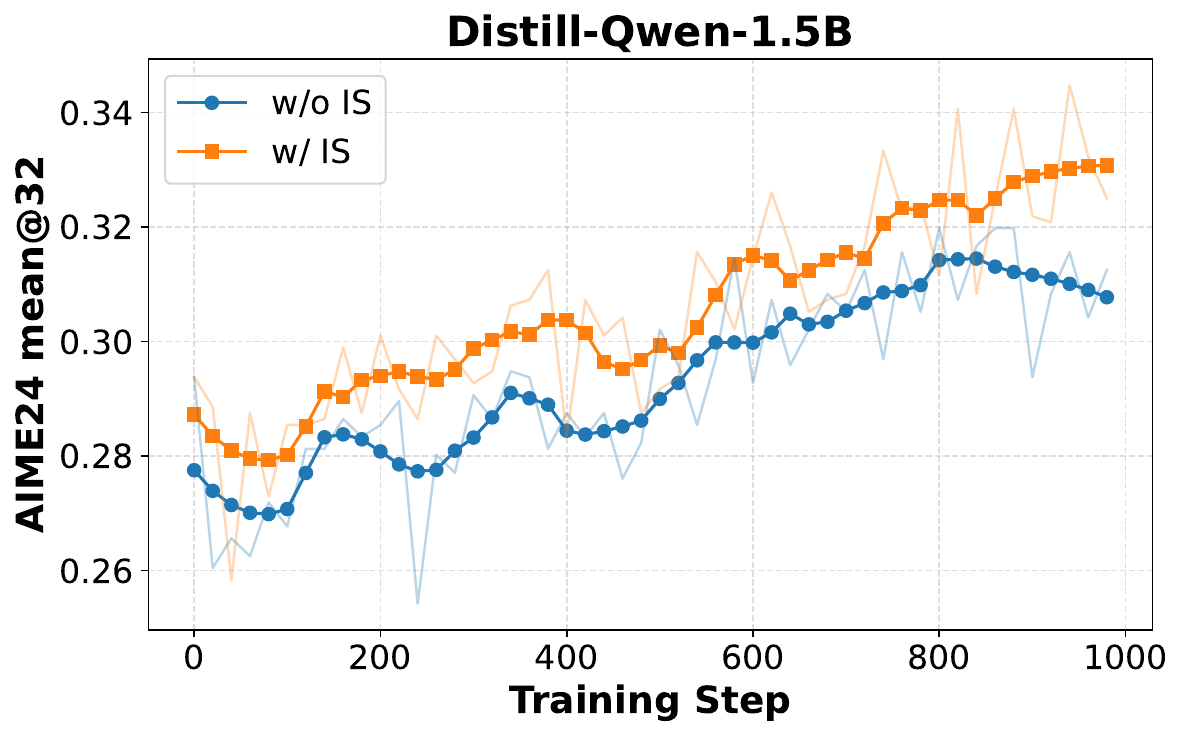}
    \label{fig:sub1}
\end{subfigure}
\hfill
\begin{subfigure}[t]{0.48\textwidth}
    \centering
    \includegraphics[width=\linewidth]{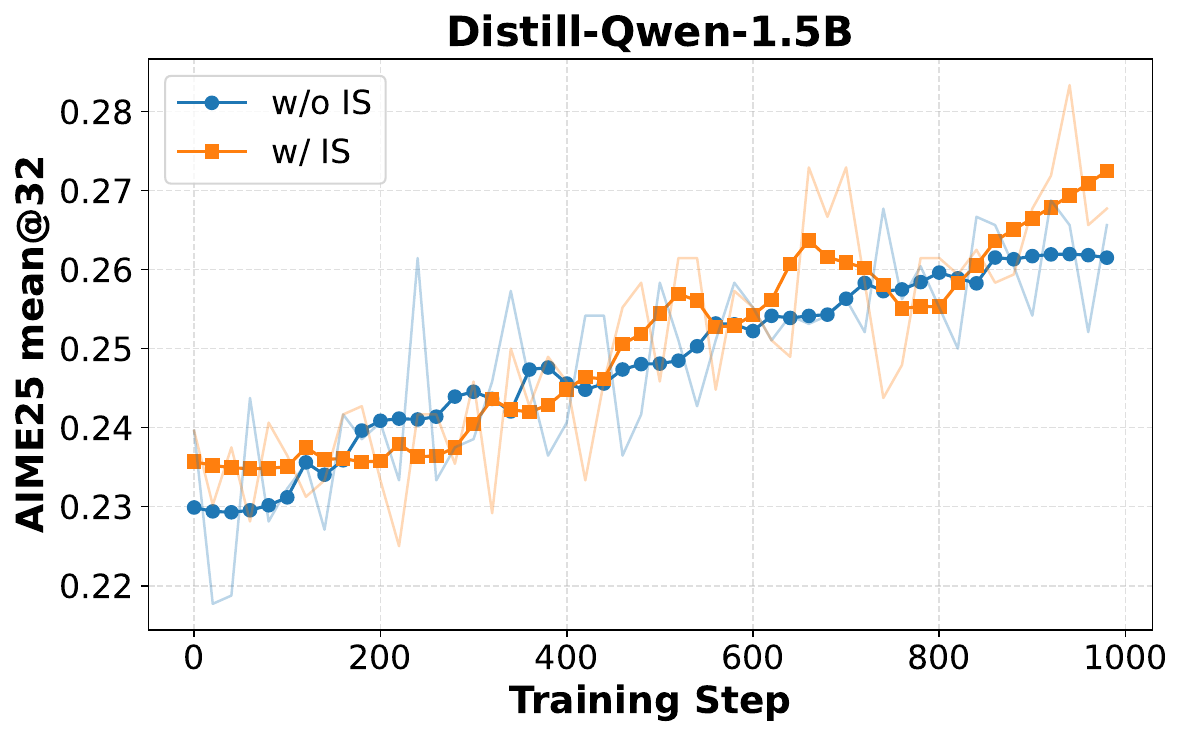}
    \label{fig:sub2}
\end{subfigure}

\vspace{-6pt} 

\begin{subfigure}[t]{0.48\textwidth}
    \centering
    \includegraphics[width=\linewidth]{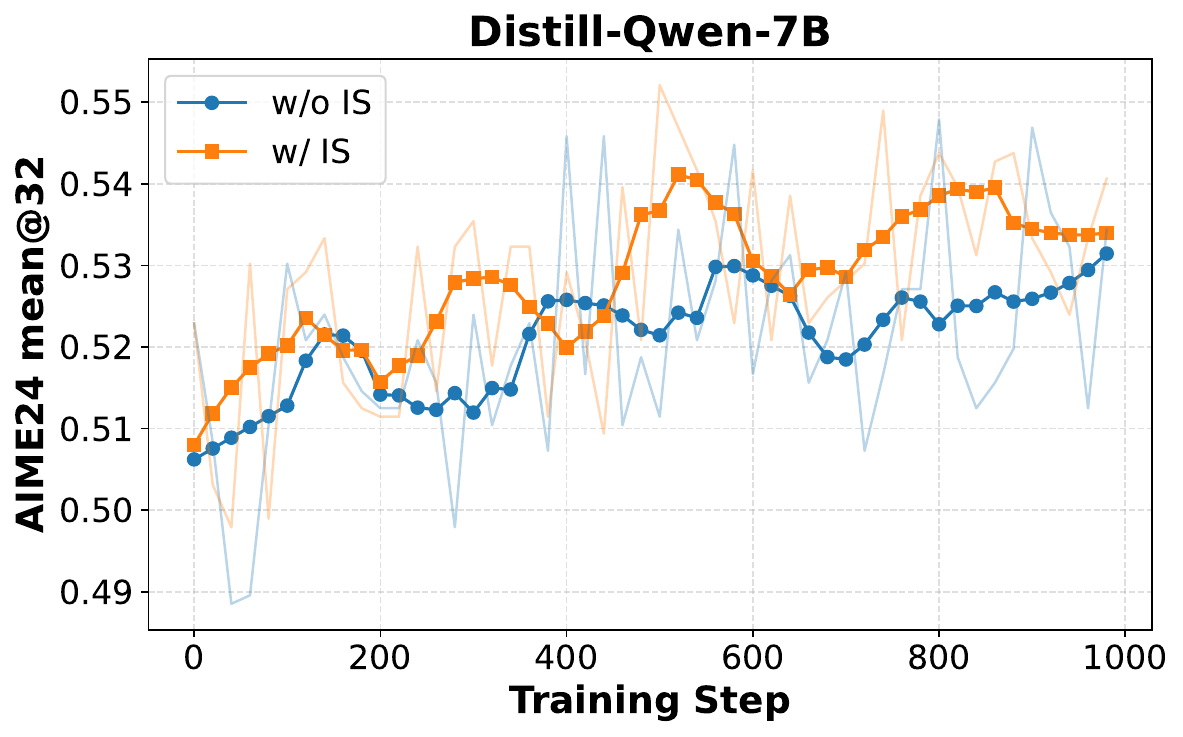}
    \label{fig:sub3}
\end{subfigure}
\hfill
\begin{subfigure}[t]{0.48\textwidth}
    \centering
    \includegraphics[width=\linewidth]{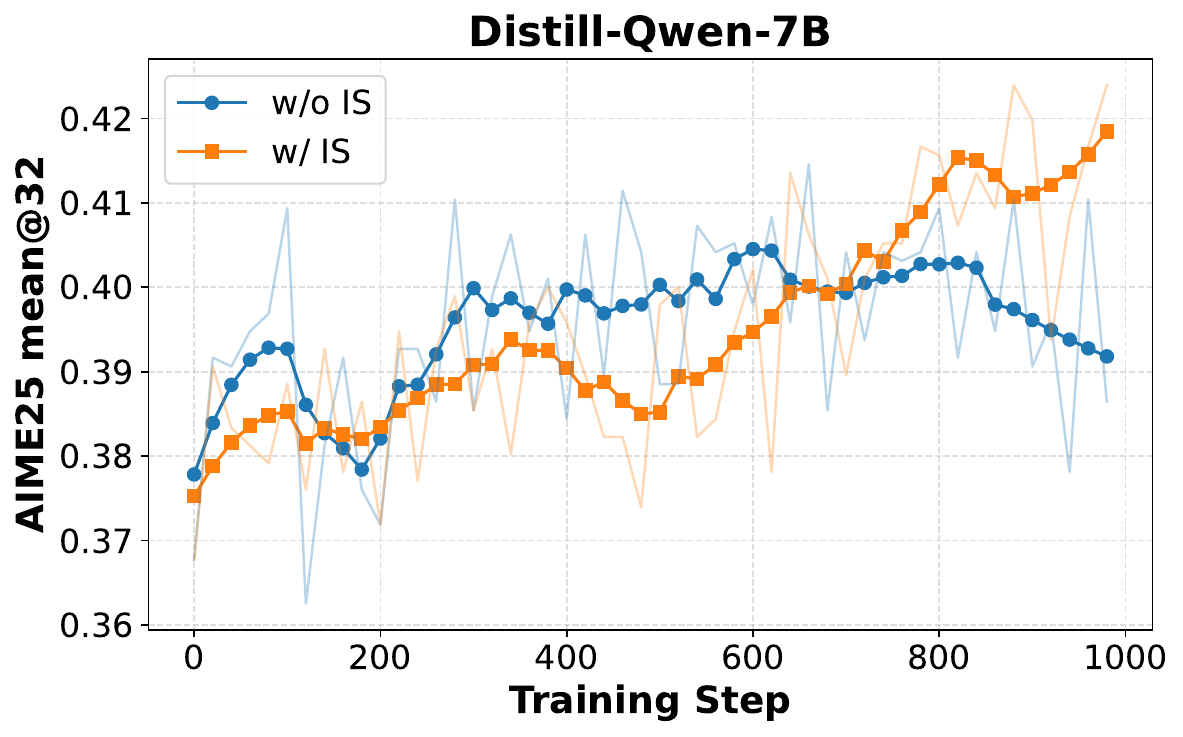}
    \label{fig:sub4}
\end{subfigure}
\vspace{-6pt} 
\caption{
Ablation results of the Cross-stage Importance Sampling Correction across two model scales. 
The top row corresponds to Distill-Qwen-1.5B and the bottom row to Distill-Qwen-7B, showing model performance on AIME24 (left) and AIME25 (right).
}
\label{fig:four_subfigs}
\end{figure*}

\subsection{Ablation Study}
To present the contribution of our proposed Concurrency-Controlled Generation and Cross-stage Importance Sampling Correction, we conduct ablation studies on these two methods.

\subsubsection{Ablation on Concurrency-Controlled Generation}
\label{ablation on concur}
\noindent
In \modelname{}, we maintain a fixed number of concurrent rollouts during the rollout stage to ensure full utilization of computational resources. This avoids the load imbalance problem in naive partial rollout, where submitting a fixed batch of requests at once causes some GPUs to handle disproportionately more short-response rollouts, leading to reduced concurrency and lower utilization in the later stage of rollout.
For inference engines, increasing concurrent requests typically improves throughput. However, excessive concurrency triggers recomputation overheads and increases off-policy trajectories requiring re-prefilling and log probability calculation for importance sampling correction. Let the concurrency level be $N'$. When collecting enough rollouts for a training batch, $N'-1$ partially generated trajectories remain in the buffer. As $N'$ grows, these off-policy trajectories accumulate, degrading training stability and throughput. Therefore, controlling concurrency is crucial for balancing efficiency and stability.

To validate Concurrency-Controlled Generation, we trained DeepSeek-R1-Distill-Qwen-1.5B on the DeepScaleR-Preview Dataset for 1000 steps under different concurrency settings and naive partial rollout with initial concurrency 1536 (matching the off-policy level of concurrency 1024 in \modelname{}). As shown in Table~\ref{tab:concurrency}:
(1) \textit{\modelname{} achieves superior resource utilization at equivalent off-policy levels.} With concurrency 1024, \modelname{} delivers a 3\% end-to-end training speedup over naive partial rollout (initial concurrency 1536) while maintaining comparable performance, demonstrating effective mitigation of load imbalance from varying rollout lengths.
(2) \textit{Moderate concurrency optimizes both efficiency and performance.} Increasing concurrency from 512 to 1024 significantly reduces training and rollout time. However, further increases raise average runtime and cause noticeable performance degradation, indicating that excessive off-policy trajectories harm training quality.

\subsubsection{Ablation on Cross-stage Importance Sampling Correction}

To verify the effectiveness of our proposed Cross-stage Importance Sampling Correction in \modelname{} training, we conducted ablation studies on both DeepSeek-R1-Distill-Qwen-1.5B and DeepSeek-R1-Distill-Qwen-7B. 
Each model is trained for 1000 steps on the DeepScaleR-Preview dataset, with results presented in Figure~\ref{fig:four_subfigs}. 
In these experiments, \textit{w/ IS} denotes our \modelname{}, which applies importance sampling correction using concatenated log probabilities. In contrast, \textit{w/o IS} refers to pseudo on-policy training that directly uses the current policy model's log probabilities without correction.
From the experimental results, we observe that:
(1) \textit{Overall effectiveness}: Models trained with Cross-stage Importance Sampling Correction consistently achieve superior performance across both model scales, demonstrating the effectiveness of our proposed correction mechanism.
(2) \textit{Scale-dependent impact}: The importance of IS correction becomes more pronounced for larger models. Specifically, the 7B model exhibits stable improvements with IS correction, while the w/o IS variant demonstrates volatile training dynamics with unstable convergence. In contrast, the 1.5B model shows relatively smaller stability and performance gaps between the two variants, though it still benefits from IS correction. We attribute this phenomenon to the inherent characteristics of model scaling: compared to the 1.5B model, the 7B model tends to produce more concentrated output distributions and exhibits larger discrepancies in log-probability distributions between policy versions across different training steps. Without proper importance sampling correction, these distributional shifts prevent the model from receiving accurate gradient signals, leading to optimization instability. These results validate the critical role of Cross-stage Importance Sampling Correction in enabling stable and efficient RL training under the \modelname{} framework, with its importance scaling with model size.

\section{Conclusion}
In this paper, we introduce \modelname{}, a simple yet effective method that mitigates the long-tail issue in RL training and enhances efficiency. By early terminating rollouts and reusing unfinished trajectories, \modelname{} eliminates idle computation and substantially boosts training throughput. With controlled concurrency and cross-stage importance sampling correction, \modelname{} achieves efficient and stable training.  Experiments demonstrate that \modelname{} delivers lossless efficiency gains to state-of-the-art RL training systems while maintaining comparable or superior performance, and scales well across context lengths and model sizes. This work takes a practical step toward building more efficient RL training frameworks, and we hope it inspires future work to design RL systems that more effectively leverage modern hardware.

\bibliography{iclr2025_conference}
\bibliographystyle{iclr2025_conference}

\newpage
\appendix
\section{Implementation Details}
\label{sec:appendix}

This section provides the hyperparameter settings and implementation details in our experiments.

\subsection{GRPO Details}
We conduct experiments using GRPO, where the reward is assigned as 1 at the final token if the generated answer is correct, and 0 otherwise. All other training hyperparameters and configurations are provided in Table~\ref{tab:setup}.

\begin{table}[t]
\centering
\setlength{\tabcolsep}{6pt}
\renewcommand{\arraystretch}{1.15}
\small
\caption{Training configurations and hyperparameters.}
\label{tab:setup}

\begin{tabularx}{0.6\textwidth}{@{}>{\RaggedRight\arraybackslash}X
                                    >{\raggedleft\arraybackslash}p{2.8cm}@{}}
\toprule
\textbf{Hyperparameter} & \textbf{Value} \\
\midrule

\multicolumn{2}{@{}l}{\textbf{Rollout Configuration}} \\[2pt]
\midrule
Rollout batch size            & 64 \\
Number of samples per prompt  & 8 \\
Rollout max prompt length     & 1024 \\
Rollout max response length   & 15360 \\
Rollout top-p                 & 1.0 \\
Rollout top-k                 & -1 \\
Rollout temperature           & 1.0 \\
Number of samples per eval prompt & 32 \\
Eval max prompt length        & 1024 \\
Eval max response length      & 31744 \\
Eval rollout temperature      & 0.6 \\
Eval rollout top-p            & 1.0 \\
Eval rollout top-k            & -1 \\

\addlinespace[3pt]
\midrule
\multicolumn{2}{@{}l}{\textbf{CoPRIS Specific Configuration}} \\[2pt]
\midrule
Concurrency pool size         & 1024 \\

\addlinespace[3pt]
\midrule
\multicolumn{2}{@{}l}{\textbf{Training Configuration}} \\[2pt]
\midrule
Global batch size             & 64 \\
Optimizer                     & Adam \\
Learning rate                 & $1\times10^{-6}$ \\
Weight decay                  & 0.01 \\
Entropy coefficient            & 0.0 \\
KL coefficient                 & 0.0 \\
KL loss type                  & low var kl \\
Clip ratio low              & 0.2 \\
Clip ratio high             & 0.28 \\
Loss aggregation mode         & token mean \\

\bottomrule
\end{tabularx}
\end{table}

\subsection{Experimental Environment}

In our experiments, we use the latest release of the veRL repository (main branch, August 18, 2025) \citep{verl} to evaluate both training throughput and model performance. The generation backend is implemented using vLLM \citep{vllm} v0.8.4, and training is conducted with PyTorch FSDP \citep{fsdp}.

\end{document}